\documentclass[letterpaper, 10 pt, conference]{ieeeconf}
\IEEEoverridecommandlockouts		
\usepackage{balance}

\usepackage{amsmath}
\usepackage{amssymb}

\usepackage{tabularx}
\usepackage{graphicx}
\usepackage{bm}
\usepackage{color}
\usepackage{float}
\usepackage{units}
\usepackage{url}
\usepackage{hyperref}

\makeatletter 
\def\endfigure{\end@float} 
\def\endtable{\end@float}
\makeatother

\newcommand{\pos}{\bm{p}}

\usepackage{subcaption}
\usepackage{caption}
\usepackage{amssymb}
\usepackage{amsmath}
\usepackage[belowskip=0pt,aboveskip=3pt]{caption}
\renewcommand{\unit}[1]{{\rm #1} }

\newcommand{\update}[1]{\textcolor{blue}{#1}}

\IEEEoverridecommandlockouts

\begin{document} 

\title{\Large \bf 			
Balancing Control and Pose Optimization for Wheel-legged Robots Navigating High Obstacles 
}

\author{Junheng Li, Junchao Ma, and Quan Nguyen 
\thanks{This work is supported by USC Viterbi School of Engineering startup funds.}
\thanks{The authors are with the Department of Aerospace and Mechanical Engineering, University of Southern California, Los Angeles, CA 90089.
email:{\tt\small junhengl@usc.edu, junchaom@usc.edu, quann@usc.edu}}%
}%
	
\maketitle

\begin{abstract}

In this paper, we propose a novel approach on controlling wheel-legged quadrupedal robots using pose optimization and force control via quadratic programming (QP). 
Our method allows the robot to leverage the whole-body motion and the wheel actuation to roll over high obstacles while keeping the wheel torques to navigate the terrain while keeping the wheel traction and balancing the robot body.
In detail, we first present a linear rigid body dynamics with wheels that can be used for real-time balancing control of wheel-legged robots.
We then introduce an effective pose optimization method for wheel-legged robot's locomotion over steep ramp and stair terrains. The pose optimization solves for optimal poses to enhance stability and enforce collision-fee constraints for the rolling motion over stair terrain.
Experimental validation on the real
robot demonstrated the capability of rolling up on a ${ 0.36\:\unit{m}}$ obstacle. The robot can also successfully roll up and down multiple stairs without lifting its legs or having collision with the terrain.

\end{abstract}


\section{Introduction}
\label{sec:Introduction}

The recent technological and theoretical developments in both robot design and controls have allowed the world to witness many successful and highly-autonomous legged robots. 
With such hardware and software advancements, researchers in the robotics field are now facing a challenge to develop mobile legged robots that can conduct given tasks fully autonomously and that the control framework can perform robustly in terrains with uneven surfaces with obstacles \cite{kajita2008legged}. 
Glancing over the development of legged robots in the last decade, many bipedal and quadruped robots have demonstrated outstanding maneuverability and dynamic locomotion in unknown terrain and have proven to have great potential to be controlled autonomously (e.g. ATRIAS \cite{nguyen2017dynamic}, Cassie \cite{gong2019feedback}, MIT Cheetah 3 \cite{bledt2018cheetah}, ANYmal quadruped \cite{hutter2016anymal}, and Boston Dynamics Spot autonomous exploration mission \cite{bouman2020autonomous}) However, energy efficiency in the robot hardware remains one of the most important condition that determines whether a mobile robot can perform real-life task that require extended period of time while maintaining highly dynamical locomotion, such as rescue and disaster-response missions \cite{seok2014design}. 
Legged robots rely on gait sequence and proper foot placement to overcome obstacles and uneven surfaces, which is a more effective method in rough terrain locomotion compared to only wheeled system \cite{pongas2007robust}, while wheeled systemss generally have far less energy consumption and faster speed to maneuver on an even surface \cite{schwarz2016hybrid}. Hence, the hybrid system of both wheels and legs could leverage the advantages from both worlds, enabling maneuverability in rough terrain, energy efficiency, and speed (e.g. \cite{grand2004stability} and \cite{grand2004decoupled}).
\update{
\begin{figure}[t]
		\center
		\includegraphics[width=0.85 \columnwidth]{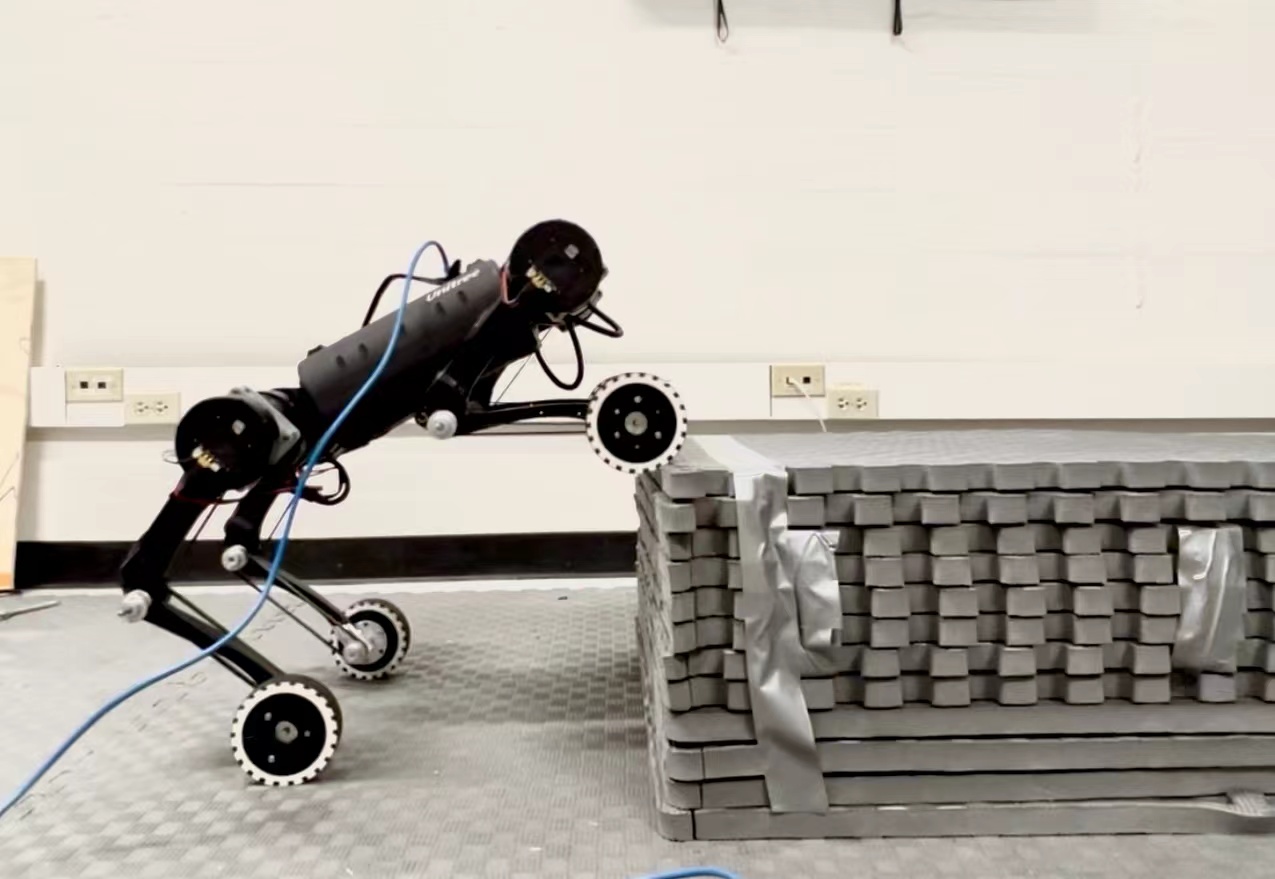}
		\caption{Snapshot of our wheel-legged robot rolling up a ${ 0.30\:\unit{m}}$ obstacle. Experiment and simulation video: \protect\url{https://youtu.be/s68-tvb1tb4}. }
		\label{fig:title}
\end{figure}
}
\subsection{Related Work}
\label{relatedWork}
Recent development in the control of wheel-legged robots shows promising results. The wheel-legged quadruped ANYmal \cite{bjelonic2020whole} utilizes a kinodyanmics model in whole-body model-predictive control for robust locomotion control. The wheel-legged bipedal robot Ascento \cite{klemm2020lqr} adopts the whole-body dynamics by using linear-quadratic regulator (LQR) and Zero-moment Point (ZMP) in balance control. 
Our approach in controlling wheel-legged robots leverages the wheel traction to traverse challenging terrains (i.e., instead of stepping over a high obstacle, we take advantage of the whole body motion and wheel actuation to roll over the obstacle).  
Note that the obstacle height a legged robot can step over is limited, our approach could allow the robot to roll over an obstacle that is higher than the robot's nominal standing height. 
This work also includes pose optimization to solve for optimal and collision-free poses for the robot rolling up high stairs or obstacles. 

Force-based QP balancing control has successful implementations on quadruped robot MIT Cheetah 3 \cite{bledt2018cheetah} and allows the robot to balance even after performing very dynamical and aerial tasks such as jumping \cite{nguyen2019optimized}. The idea of modifying simplified dynamics in QP balance controller has also gained success in controlling mobile legged robots in the past (e.g. \cite{li2021force}). 
In our work, we develop this control paradigm for wheel-legged robots by considering the wheel dynamics and terrain slope in our model. We also adopt this for the rolling task instead of just standing balance.

Trajectory optimization frameworks in motion planning have allowed legged robots to achieve highly-dynamic locomotion. For example, \cite{katz2019mini}, \cite{nguyen2019optimized}, \cite{bellegarda2019trajectory}, and \cite{winkler2018gait} all utilize NLP solver to find optimal trajectories for a certain task. One important constraint in such optimization frameworks is the utilization of robot dynamics, either full body dynamics or simplified centroidal dynamics. In \cite{luo2014design}, a NLP parameter optimization is used to find the optimal driving force in achieving highly dynamic locomotion. 
In wheel-legged robot's motion planning, our approach with pose optimization 
uses kinematic constraints instead to guarantee collision-free with the terrain and solve for favorable configurations to maintain balance.
The advantage of this framework is that the robot can leverage and adapt to the obstacle's shape and height to overcome it rather than simply stepping or rolling over it. Thus, it allows the robot to overcome an obstacle with a height that exceeds its nominal standing height. 
In our pose optimization framework, we choose not to use dynamics or Inverse Kinematics (IK) to solve for joint angles by relative foot position (i.e. kinodynamics models employed in \cite{bjelonic2020whole}, \cite{medeiros2020trajectory}, and \cite{chignoli2021humanoid}), instead, we directly use joint angles, body center of mass (CoM) location in 2D, and body pitch angle as the only optimization variables and use Forward Kinematics (FK) to constrain the relative foot position and collision-avoidance in a favorable pose, which allows us to have a much faster solving time. We consider the dynamics of the robot in our force-based feedback controller for real-time motion planning and control. 
The optimal pose will then be used in combination with a balance controller using QP-based force control to maintain balance and desired pitch angle. 
Pose optimization also happens to resonate with the crawling mode introduced in \cite{besseron2005locomotion} and \cite{amar2004performance}, in which the crawling mode in rolling is proven to be a superior option in slope-climbing. Thanks to only few critical poses being needed, the computation intensity is dramatically scaled-down compared to full trajectory optimizations. Unlike the approach in  \cite{grand2004decoupled} by adapting wheel-legged robot's posture in rough terrain with feedback control, or the approach in \cite{ni2020posture} by adding passive suspension for pose adapting, our approach of finding the optimal poses or robot configurations based on the terrain map.

\subsection{Contributions}
\label{contributions}

The main contributions of the paper are as follows:
\begin{itemize}
    \item 
    We introduce a new rigid body dynamics with wheels dynamics that can be effectively used for force-based balancing control of wheel-legged robots.
    \item Our proposed pose optimization method with kinematic and collision-free constraints only requires solving very few critical poses in a task that consists of high obstacles while maintaining wheel traction with the terrain. (only 2 poses are needed to solve in a single-stair task)
    
    \item The pose optimization is very efficient due to its small problem size. 
    The solved optimal poses at a certain location can be linearly interpolated to obtain the joint trajectory at any given time during the task.
    \item We use a hybrid control framework that includes force-based QP and joint PD control to track optimal poses in order to achieve stable locomotion of wheel-legged robots navigating terrain with high obstacles.
    \item Experimental validation on the real robot demonstrated the capability of rolling up on a ${ 0.36\:\unit{m}}$ obstacle. The robot can also successfully roll up and down multiple stairs without lifting its legs or having collision with the terrain.

\end{itemize}

The rest of the paper is organized as follows. Section \ref{sec:robotModel} introduces the wheel-legged robot model, its physical parameters. 
The overview of our control architecture is highlighted in Section \ref{sec:overview}.
Section \ref{sec:dynamicsAndControl} presents the rigid body dynamics with wheels used in balancing control for wheel-legged robots, force-based QP control formulation, and rolling control. 
Section \ref{sec:PoseOptimization} introduces the pose optimization and its problem formulations, constraints, and real-time pose planning. Section \ref{sec:Results} highlights some experimental results with the proposed approach.  


\section{{Robot Model and Hardware}}
\label{sec:robotModel}

\begin{figure}[h]
		\center
		\includegraphics[width=0.55 \columnwidth]{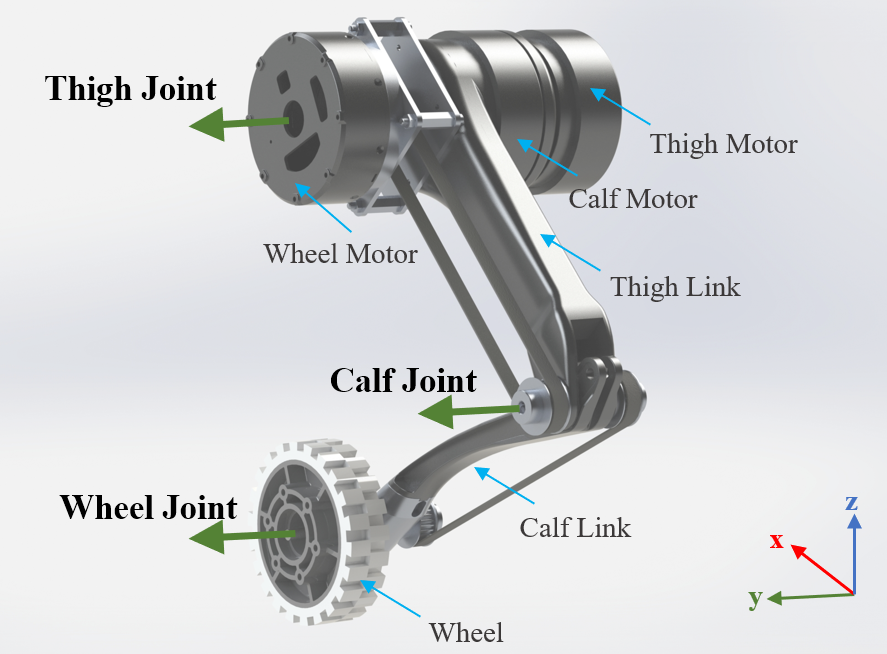}
		\caption{{\bfseries Leg Configuration.} The link and joint configuration of wheel-legged robot leg, rendered in SolidWorks.}
		\label{fig:legConfig}
	\end{figure}

In this section, we introduce the model of our wheel-legged robot and its physical parameters.
Fig. \ref{fig:title} presents the hardware assembly of the wheel-legged robot in motion.
Our wheel-legged robot consists of a robot trunk, four sets of 3 degrees of freedom(DoF) legs.
Each leg consists of the thigh, calf, and wheel. In total, the wheel-legged robot has 12 actuators. Fig. \ref{fig:legConfig} shows the leg assembly and joint definitions. 
Table \ref{tab:PRP} includes all physical parameters of the robot.
Our robot is built from the Unitree A1 quadruped robot with Unitree A1 actuators. The A1 actuator is a torque-controller motor that can provide 21.0 $\unit{rad/s}$ maximum angular speed and 33.5 $\unit{Nm}$ maximum torque output. 


\begin{table}[h]
	\hspace{0.2cm}
	\centering
	\caption{Robot Physical Parameters}
	\label{tab:PRP}
	\begin{tabular}{cccc}
		\hline
		Parameter & Symbol & Value & Units\\
		\hline
		Mass & $\bm m$    & 11.84 & $\unit{kg}$  \\[.5ex]
		Body Inertia  & $I_{xx}$  & 0.0214 & $\unit{kg}\cdot \unit{m}^2$ \\[.5ex]
		& $I_{yy}$ & 0.0535  & $\unit{kg}\cdot \unit{m}^2$ \\[.5ex]
		& $I_{zz}$ & 0.0443  & $\unit{kg}\cdot \unit{m} ^2$ \\[.5ex]
		Body Length & $l_{b}$ & 0.247 & $\unit{m}$ \\[.5ex]
		Body Width & $w_{b}$ & 0.194 & $\unit{m}$ \\[.5ex]
		Body Height & $h_{b}$ & 0.114 & $\unit{m}$ \\[.5ex]
		Thigh and Calf Lengths & $l_{1}, l_{2}$ & 0.2 & $\unit{m}$ \\[.5ex]
		Wheel Radius & $\bm R_{wheel}$ & 0.05 & $\unit{m}$ \\[.5ex]
		\hline 
		\label{tab:robot}
	\end{tabular}
\end{table}

\section{Overview}
\label{sec:overview}

\begin{figure}[h]
	\hspace{0.2cm}
		\center
		\includegraphics[width=1 \columnwidth]{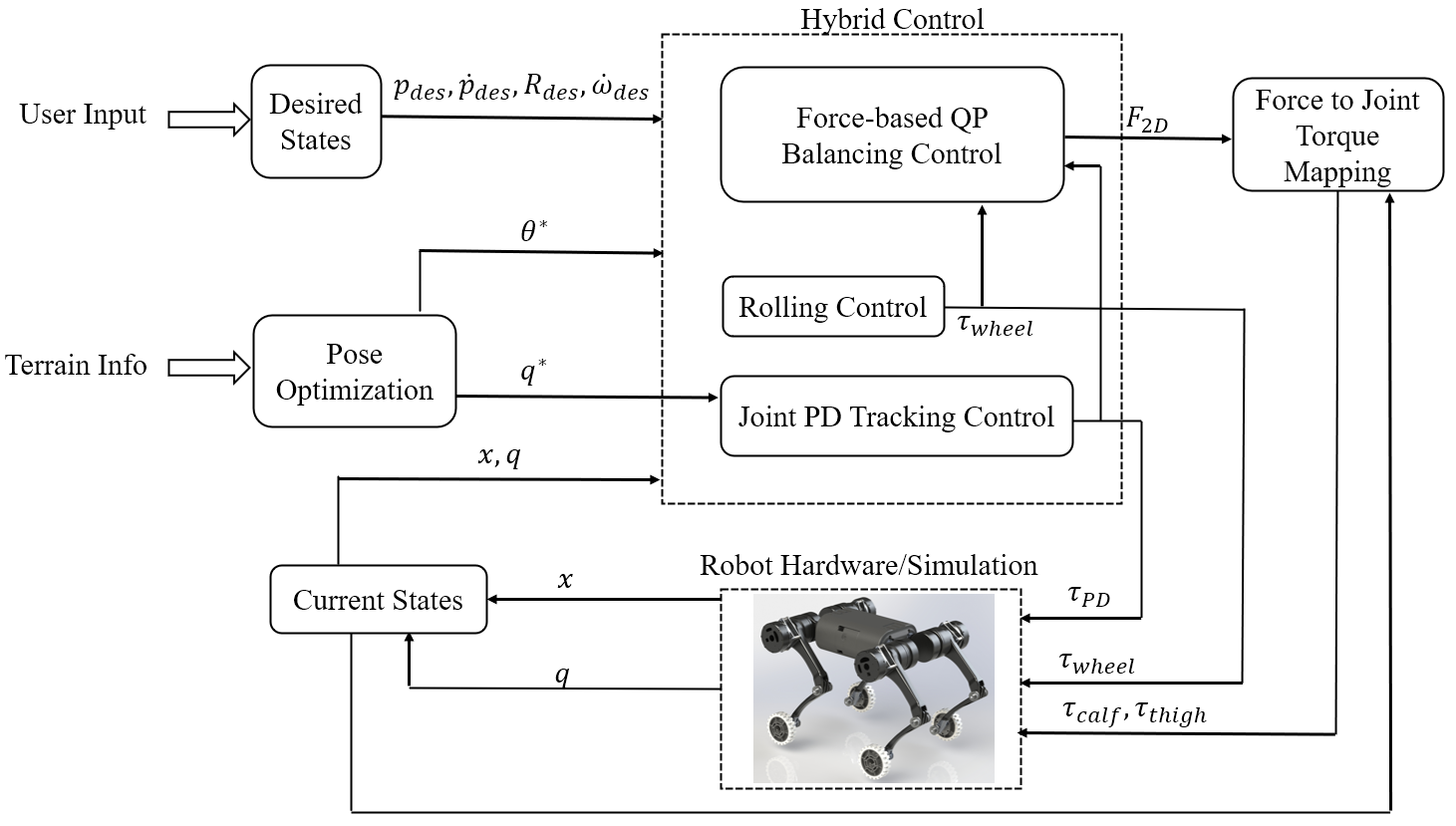}
		\caption{{\bfseries Control Architecture.} The block diagram for our proposed approach. }
		\label{fig:controlArchi}
	\end{figure}

Having presented the robot model and hardware, in this Section, we highlight the overview of our control architecture and critical parts of our proposed framework.
The control architecture and block diagram are present in Fig. \ref{fig:controlArchi}. 
We use QP force-based balance controller to maintain balance in various tasks. This balancing control only commands the thigh and calf joint torques. For controlling the wheels, the rolling control enables the robot to maneuver with wheel traction and yaw on command. 
The details of the balancing control will be explained in Section \ref{sec:dynamicsAndControl}. 
In order to control our robot to roll over challenging terrains (e.g., stairs, high obstacles, or steep ramps), we also propose a pose optimization framework to solve for optimal configuration of the robot that is collision-free with the terrain while maintaining a good support region for the robot to keep the body balanced.
Section \ref{sec:PoseOptimization} will explain in more detail this method.
The desired pitch angle is fed into the balance controller to maintain balance during motion. And the joint angles are tracked by joint PD control to manipulate the robot pose.

\section{Force-based Balance Control}
\label{sec:dynamicsAndControl}
\subsection{Rigid-body Dynamics with Wheels}
\label{subsec:simplified_dynamics}
In this section, we introduce a simplified dynamics model for wheel-legged robots that can be used effectively in real-time feedback control for balancing the robot body. 
Single rigid-body dynamics are commonly used for controlling quadruped robots \cite{focchi2017high,nguyen2019optimized}.
However, when the robot highly leverages the wheels in traversing uneven terrain, it is critical to take into account the impact of wheel traction forces on the robot dynamics. In our work, the simplified dynamics model takes wheel dynamics into consideration. 
Our control framework has been divided into balancing control and rolling control. The balancing controller output is mapped to only the thigh and calf torques, and the wheel torque is controlled by the rolling controller.
The two decoupled control methods are connected by combining the wheel dynamics (Fig. \ref{fig:wheelDynamics}) with the simplified rigid body dynamics of the quadruped robot to become a hybrid linear dynamics model (Fig. \ref{fig:SRBD}), which is to be used in QP balancing controller. Considering the wheel dynamics in angular motion while on the ground, shown in Fig. \ref{fig:wheelDynamics}:
\begin{figure}[!h]
     \center
     \begin{subfigure}[b]{0.16\textwidth}
         \centering
         \includegraphics[width=\textwidth]{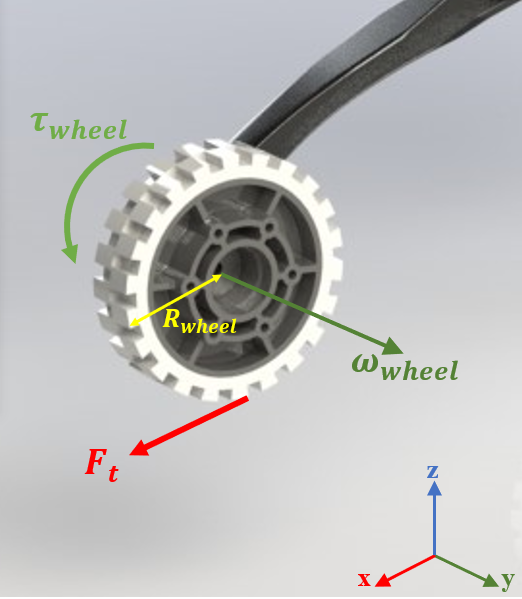}
         \caption{Wheel Dynamics}
         \label{fig:wheelDynamics}
     \end{subfigure}
     \begin{subfigure}[b]{0.31\textwidth}
         \centering
         \includegraphics[width=\textwidth]{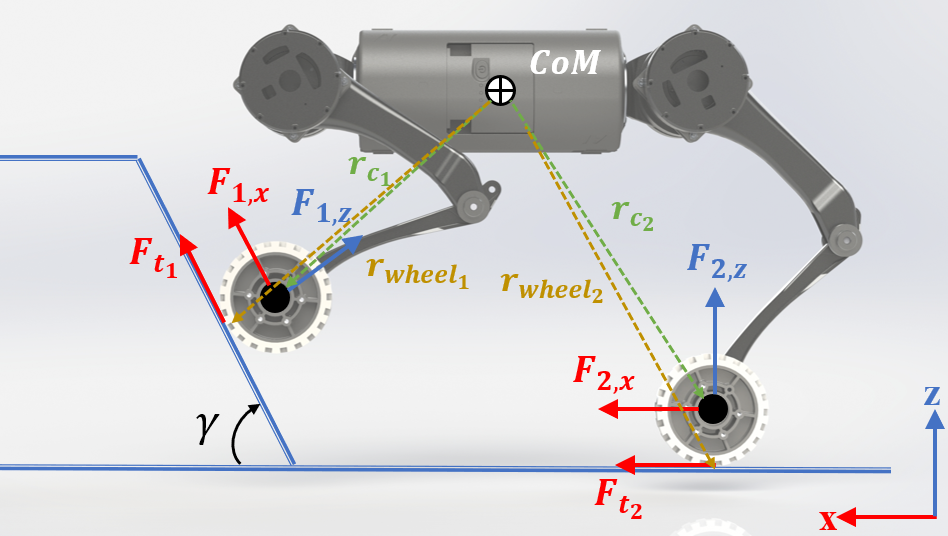}
         \caption{2D Simplified Rigid Body Dynamics with Wheel }
         \label{fig:SRBD}
     \end{subfigure}

        \caption{{\bfseries Robot Dynamics} Wheel-legged Robot Dynamics} 
        \label{fig:Dynamics}
\end{figure}
\begin{align}
\label{eq:wheelDynamicsFull}
\bm I_{wheel}\cdot\dot{\bm{\omega}}_{wheel} = \bm{\tau}_{{wheel}_i}-\bm F_{t_i}\cdot\bm R_{wheel},
\end{align} 
where $\bm I_{wheel}$ is the rotation inertia of the wheel, $\dot{\bm{\omega}}_{wheel}$ is the angular acceleration of the wheel, $\bm{\tau}_{{wheel}_i}$ is the wheel torque of the $i^{th}$ leg from control input, and $\bm F_{t_i}$ is the wheel traction force of the $i^{th}$ leg at ground contact point on the rim of the wheel. We choose to neglect the very small rotation inertia value of the wheel dynamics to obtain a linear relation of the wheel torque and traction force, equation (\ref{eq:wheelDynamicsFull}) is then reduced to:
\begin{align}
\label{eq:wheelDynamicsSimple}
\bm F_{t_i} = \frac{\bm{\tau}_{{wheel}_i}}{\bm R_{wheel}}.
\end{align} 

When the robot's wheel contact point is on an aggressive slope, it is important to take into consideration of the change in coordinate frames of ground reaction forces as well as friction constraints. Fig. \ref{fig:SRBD} illustrate one scenario when the front legs of the robot are on the slope with a positive angle $\gamma$, notice that the ground reaction force in z-direction has to be normal to the terrain while the ground reaction force in the x-direction has to be parallel to the terrain.
The simplified dynamics of the robot can be written as: 
\begin{align}
\label{eq:simplifiedDynamics}
\bm A_c\cdot \bm F + \bm A_{wheel} \cdot \bm F_{wheel}=\bm b, 
\end{align} 
where 
\begin{align}
\label{eq:Ac}
\bm A_c= 
\begin{bmatrix}
\bm R_{\gamma} & \dotsc  & \mathbf I_{2}\\
(\bm R_{\gamma}^T\bm r_{c_1}) \times  & \dotsc  & \bm r_{c_4} \times 
\end{bmatrix}
\end{align} 
\begin{align}
\label{eq:Awheel}
\bm A_{wheel}= 
\begin{bmatrix}
\bm R_{\gamma} & \dotsc  & \mathbf I_{2}\\
(\bm R_{\gamma}^T\bm r_{wheel_1}) \times  & \dotsc  & \bm r_{wheel_4} \times 
\end{bmatrix}
\end{align} 
\begin{align}
\label{eq:b}
\bm b = \left[\begin{array}{c}  m (\ddot{\pos}_{c} +\bm{g}) \\ \bm I_w\dot{\bm {\omega}} \end{array} \right].
\end{align} 
$\bm R_{\gamma}$ is 2D rotation matrix by the slope angle $\gamma$,
\begin{align}
\label{eq:RGamma}
\bm R_{\gamma} = \begin{bmatrix}
\cos( \gamma ) & \sin( \gamma )\\
-\sin( \gamma ) & \cos( \gamma )
\end{bmatrix}  
\end{align} 
The term $\bm F$ is a force vector containing 2D ground reaction forces at the center of each wheel, $\bm F = [\bm F_{1,x}, \bm F_{1,z}, \dotsc \bm F_{4,x}, \bm F_{4,z}]^T$. Similarly, $\bm F_{wheel}$ is a column vector containing the wheel traction forces obtained from equation (\ref{eq:wheelDynamicsSimple}), $\bm F_{wheel} = [\bm F_{t_1}, 0, \dotsc \bm F_{t_4}, 0]^T$. Note that the wheel only contributes force in the direction of the ground. 
In equation (\ref{eq:Ac}) and (\ref{eq:Awheel}), $\bm r_{c_i}$ is the vector of distance from trunk CoM to center of the $i^{th}$ wheel and $\bm r_{wheel_i}$ is the distance from CoM to ground contact point of the $i^{th}$ wheel, $i=1,\dotsc,4$,
 $\bm r_{c_i}\times = [\bm r_{{c_i},z},  -\bm r_{{c_i},x}]$, and $\bm r_{{wheel}_i}\times = [\bm r_{{{wheel}_i},z},  -\bm r_{{{wheel}_i},x}]$. 
%
In equation (\ref{eq:b}), $\ddot{\pos}_{c}$ is the linear acceleration of the robot CoM in 2D (x and z-direction), $\bm g$ is the gravity vector in 2D, $\bm I_w$ is the rotation inertia of the robot body in the world frame, and $\dot{\bm {\omega}}$ is the angular acceleration of the robot body around y-axis. 

Similarly, if the robot's rear wheels are in contact with a sloped surface, the formulation in equation(\ref{eq:Ac}) and (\ref{eq:Awheel}) should be modified to reflect the change of frames of the corresponding ground reaction forces.

\subsection{QP Formulation of Balance Controller}
\label{subsec:qpBalance}
Since the dynamics model \ref{eq:simplifiedDynamics} presented in the previous section is linear, we can incorporate the dynamics constraints in a quadratic program (QP) as follow. The formulation of this balance controller is inspired by \cite{focchi2017high} which was developed for quadruped robots using a single rigid body dynamics. In this work, we adopt this principle using our model of rigid body dynamics with wheels for our wheel-legged robots.
The balancing control employs a PD control policy of the robot body CoM position. It also makes sure the inequality constraints such as force saturation and friction constraints are stratified in the optimal solution. 

In this work, we will design a controller that tends to drive the robot dynamics to the following desired dynamics that follows a PD control law:
\begin{align}
\label{eq:PDlaw}
\begin{bmatrix}
\ddot{\bm p}_{c,des}{}\\
\dot{\bm \omega }_{des}
\end{bmatrix} \ \ =\ \begin{bmatrix}
\bm K_{p,p}(\bm p_{c,des} -\bm p_{c}) +\bm K_{d,p}(\dot{\bm p}_{c,des} -\dot{\bm p}_{c})\\
\bm K_{p,\omega }(\bm \theta_{des} -\bm \theta ) +\bm K_{d,\omega }(\bm \omega_{des} -\bm \omega )
\end{bmatrix}. \ 
\end{align} 
The right-hand side of equation (\ref{eq:PDlaw}) contains user input command in terms of desired CoM position, velocity, pitch angle $\bm {\theta}_{des}$ and angular velocity. The left-hand side can then be used to represent the desired $\bm b$ matrix in the dynamics equation (\ref{eq:simplifiedDynamics}):
\begin{align}
\label{eq:bdes}
\bm b_{des} = \left[\begin{array}{c} \bm m (\ddot{\pos}_{c, des} +\bm{g}) \\ \bm I_w\dot{\bm {\omega}}_{des} \end{array} \right].
\end{align} 
Then, we can obtain the desired dynamics by driving the left-hand side of the dynamics equation \eqref{eq:simplifiedDynamics} to: 
\begin{align}
\label{eq:QPdynamics}
\bm A_c\cdot \bm F \to \bm b_{des} - \bm A_{wheel} \cdot \bm F_{wheel}
\end{align} 
where the value of $\bm F_{wheel}$ is dependent on wheel torques, controller by rolling controller in Section \ref{subsec:rollingControl}.

Equation (\ref{eq:QPdynamics}) can be obtained by the following quadratic program:
\begin{align}
\label{eq:QPformulation}
\bm F_{opt} = \underset{\bm {F\in {\mathbb{R} }^8}}{\operatorname{min}}\ \bm D^T \bm S \bm D + {\alpha} \bm \parallel	\bm F\parallel^2+\beta \parallel \bm F - \bm F_{opt,prev} \parallel	^2
\end{align}
\begin{align}
\label{eq:QPconstraint}
\:\:\mbox{s.t. }& \quad  \bm C \bm F \leq \bm d
\end{align}
where, $\bm D = \bm A_c\bm F + \bm A_{wheel}\bm F_{wheel}-\bm b_{des}$.
Equation (\ref{eq:QPformulation}) is the cost function of this QP problem, its main goals are driving robot CoM location close to the desired command, minimizing the optimal force $\bm F_{opt}$, and filtering the difference of optimal force at the current time step and previous optimal force $\bm F_{opt,prev}$. These three tasks are weight by $\bm S$, $\alpha$, and $\beta$ to determine the task priorities. 
Equation (\ref{eq:QPconstraint}) summarizes the friction cone constraint and saturation of computed ground reaction force.

The resulting optimal force inputs from the QP problem in equation (\ref{eq:QPformulation}) and (\ref{eq:QPconstraint}), $\bm F_{opt} = [\bm F_{1,x}, \bm F_{1,z}, \dotsc \bm F_{4,x}, \bm F_{4,z}]^T$ are then mapped to the thigh and calf joint torques for each leg by:
\begin{align}
\label{eq:forceTorquemap}
\bm {\tau}_{QP,i} =  {\bm J_i}^T 
\begin{bmatrix}
\bm F_{i,x}\\
\bm F_{i,z}
\end{bmatrix},
\end{align}
where $\bm J_i$ is the leg Jacobian matrix of $i$th leg.


\subsection{3D Wheel Rolling Control}
\label{subsec:rollingControl}

 While the QP force control provides balance and stability to the wheel-legged robot during motion, the forward velocity and yaw control of the robot can be realized by leveraging the rolling motion of the wheels. 
 With a given CoM velocity command, the wheel torque is calculated using the following feedback law:
\begin{align}
\label{eq:rollingControl}
\bm {\tau}_{wheel} = \bm K_{d,wheel}(\dot{\bm q}_{wheel, des}-\dot{\bm q}_{wheel}),
\end{align} 
where $\dot{\bm q}_{wheel}$ is the measurement of the wheel joint angular velocity, and
\begin{align}
\label{eq:qdot}
\dot{\bm q}_{wheel, des} = \frac{\dot{\bm p}_{c_x,des}}{\bm R_{wheel}},
\end{align}
with $\dot{\bm p}_{c_x,des}$ being the desired forward velocity.
On top of this rolling control based on the input linear velocity command, the controller can also track a desired yaw speed command during rolling motion. This is achieved by assigning a difference ${\Delta}{\dot{\bm q}_{wheel, des}}$ in commanded angular speed to the left and right wheel joints, to achieve a feedback turning control.
And ${\Delta}{\dot{\bm q}_{wheel}}$ is adjusted by a yaw-speed ($\dot{\psi}$) controller,
\begin{align}
\label{eq:yawControl}
{\Delta}{\dot{\bm q}_{wheel}} = \bm K_{d,\dot{\psi}}({\dot{\psi}}_{des}-\dot{\psi}).
\end{align}
The combination of QP force-based balance control and rolling control allows the wheel-legged robot to have stable dynamic locomotion over uneven terrain by taking the advantage of wheel rolling traction.

\section{Pose Optimization}
\label{sec:PoseOptimization}

\begin{figure}[h]
	\center
	\includegraphics[width=0.9 \columnwidth]{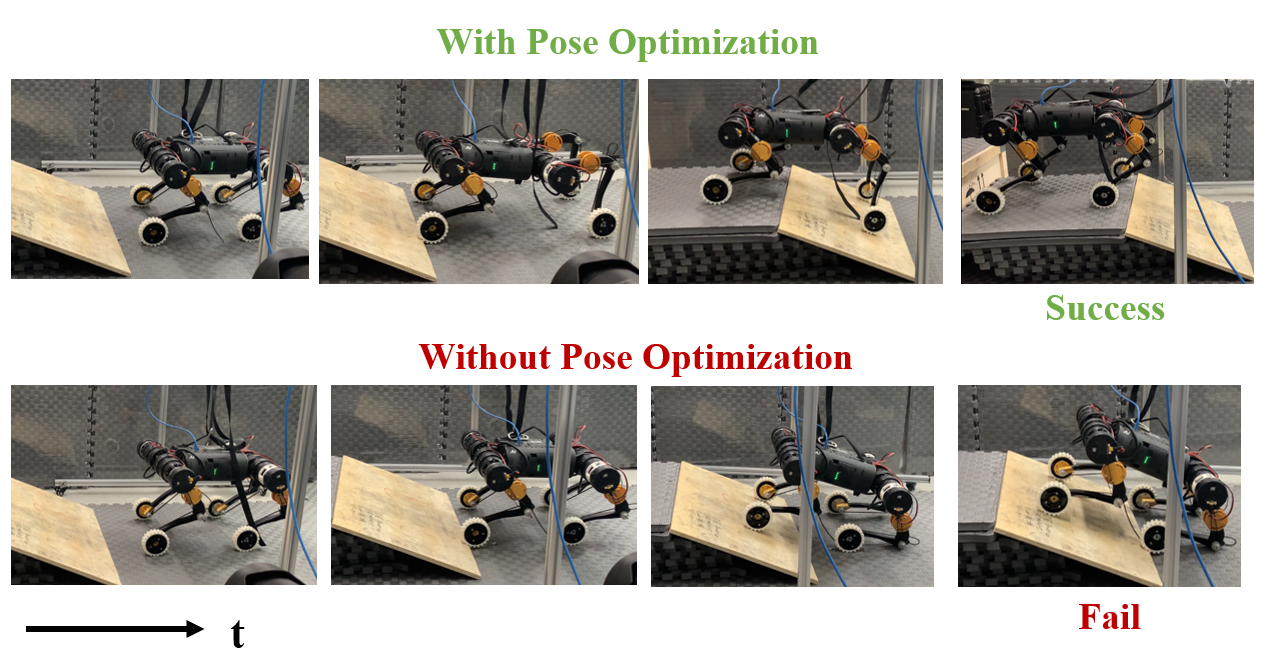}
	\caption{{\bfseries Wheel-legged Robot Rolling over a Ramp.} Experiment snapshots of rolling over a ramp with a height of 0.25$\unit {m}$ and slope angle of ${30^{\circ}}$, with and without pose optimization. }
	\label{fig:poseOptHardware}
\end{figure}

In the previous sections, we have explained the architecture of the hybrid control method that enables stable locomotion of wheel-legged robots only leveraging the wheel rolling motion.
However, with only balancing control and wheel rolling control, the robot is unable to pass more complex terrains such as terrain with a very steep slope and tall staircases, such as an example shown in Fig. \ref{fig:poseOptHardware}. 
We propose a pose optimization method based on robot kinematics and can solve for optimal poses for a given task that consists of rolling over high obstacles while ensuring collision-free constraints with the terrain. Pose optimization is a motion planning technique that can be used with terrain map information as input. This approach will allow the robot to roll over tall obstacles that exceed the robot's nominal height.

\subsection{Problem Description}
\label{subsec:problemFormulation}
\begin{figure}
	\hspace{0.2cm}
     \center
     \begin{subfigure}[b]{0.47\textwidth}
         \centering
         \includegraphics[width=\textwidth]{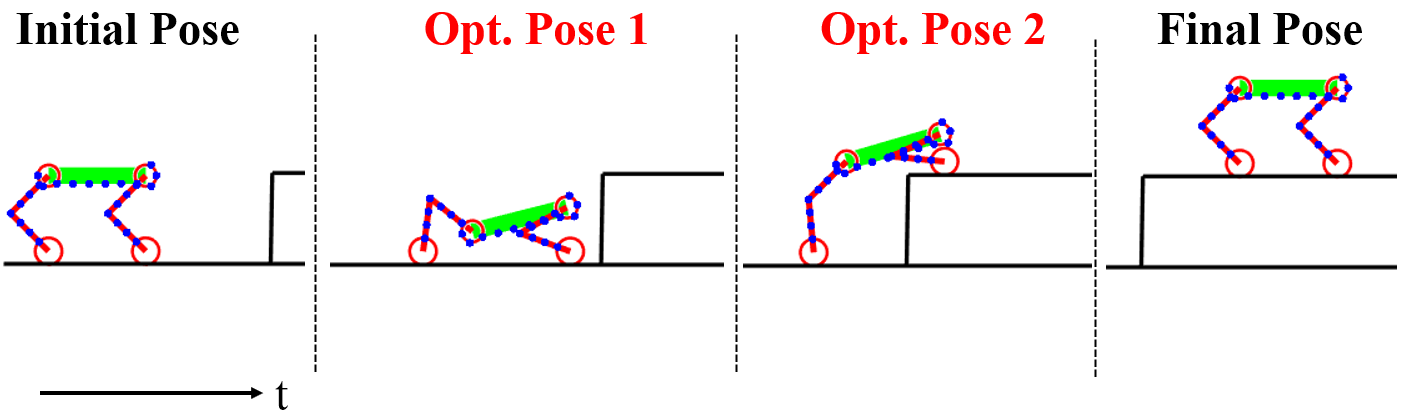}
         \caption{Four main poses during a single-stair terrain problem. The initial and final poses are known. Pose 1 and Pose 2 are solved by optimization.}
         \label{fig:poses}
     \end{subfigure}
     \\
     \begin{subfigure}[b]{0.47\textwidth}
         \centering
         \includegraphics[width=\textwidth]{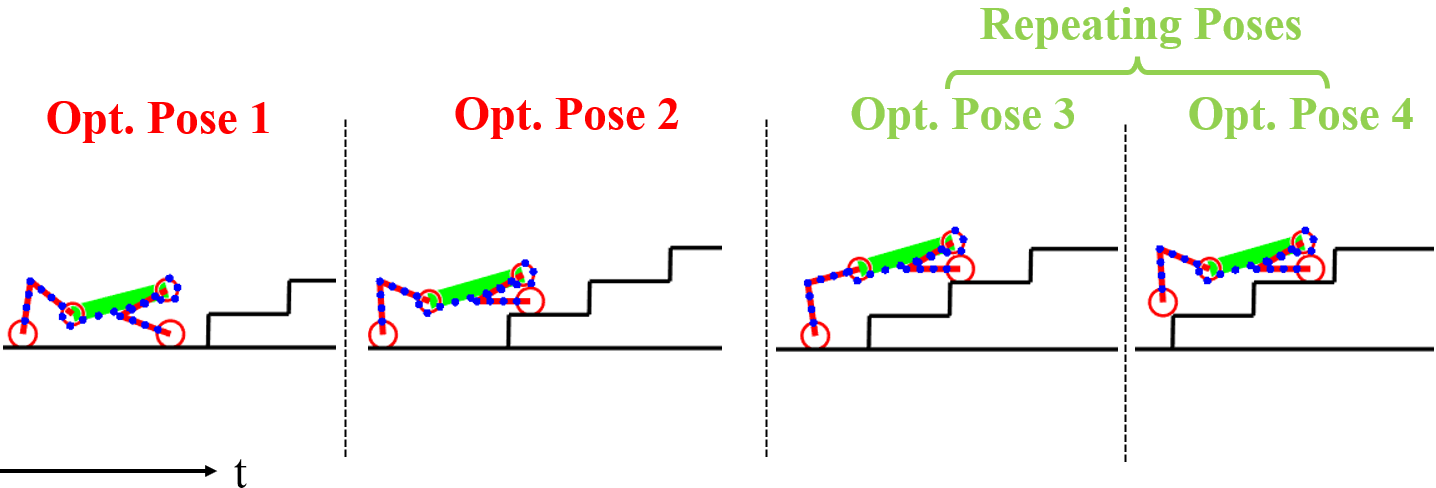}
         \caption{Optimal pose sequence for a multi-stair terrain problem. Pose 3 and 4 are repeated for each additional stair, assuming stairs have uniform rise and run.}
         \label{fig:poses_multi}
     \end{subfigure}

        \caption{{\bfseries Pose Optimization Result Illustrations} (a) Single-stair Task. (b) Multiple-stair task. The collision model is represented in blue dots in illustrations}
\end{figure}

Our approach, in the task where the wheel-legged robot needs to climb up a high stair, focuses on manipulating the robot to maintain and transform between different poses in order to create a large stability region while the robot is on a slope and while the body is at a significant pitch angle. 
In order to decrease the problem computation expense, the pose optimization only needs to compute two optimal poses at certain positions in a single-stair obstacle task. As Fig. \ref{fig:poses} has shown, the two pose locations have the largest kinematic changes during this transition from ground to the upper surface. To ensure these critical poses are collision-free, we simplified the collision model of the robot to 25 possible collision points placed across the robot trunk and limbs. The collision model is also  illustrated in Fig. \ref{fig:poses} and \ref{fig:poses_multi}. The total number of the points is determined by trial-and-error from simulation results and is the middle ground of a well-constrained problem vs. computation time. 

The optimization method also has great potential of extending its usage to more complex terrains such as multiple-stair obstacles. Two additional poses are added to the pose sequence with every additional stair. When the multi-stair terrain has uniformed stair runs and rises, the additional poses can be repeated for each additional stair without the requirement to solve repetitively, illustrated in Fig. \ref{fig:poses_multi}.

\subsection{Nonlinear Programming Problem (NLP) Formulation}
\label{subsec:Cost}
The formal NLP of the pose optimization is defined as follows,
\begin{align}
\label{eq:NLPCost}
\textit{Minimize}:\quad \bm J_i = (\bm {\bm p}_{c}^{ref}-\bm {\bm p}_{c})^T \bm Q \bm (\bm {\bm p}_{c}^{ref}-\bm {\bm p}_{c}) \\
\textit{Find}:\quad \mathbf{X}_i = [\bm x_c,\ \bm z_c,\ \bm \theta,\ \mathbf q]_i^T \quad \quad \quad \quad \:
\end{align} 
\begin{align}
\label{eq:Constraint1}
\textit{Subject to}:\quad \bm p_{wheel}(\mathbf{X}_i) = \texttt{terrain}(\bm p_c) \quad \quad \:\:\:
\end{align}
\begin{align}
\label{eq:Constraint2}
\bm p_{rw,x}(\mathbf{X}_i) \leq p_{rh,x}(\mathbf{X}_i)
\end{align} 
\begin{align}
\label{eq:Constraint3}
\texttt{InCollision}(\bm p_{cm},\texttt{terrain}(\bm p_c)) = False
\end{align} 
\begin{align}
\label{eq:Constraint4}
\mathbf q_{min}<\mathbf q <\mathbf q_{max}
\end{align} 

The objective of the pose optimization is to find the optimal pose $i$ at pose location $\bm {\bm p}_{c}^{ref}=[\bm x_c^{ref},\ \bm z_c^{ref}]^T$, the reference pose location is resulted from the terrain information. Hence, the cost function $\bm J_i$ aims to find the closest possible location that satisfy the given NLP constraints. 
In this optimization framework, we use kinematic constraints, therefore, a feasible pose solution $\mathbf X_i$ should contain CoM 2D locations $\bm x_c$ and $\bm z_c$, body pitch angle $\bm \theta$, and limb joint angles $\mathbf q^* = [\bm q_1,\ \bm q_2,\ \bm q_3,\ \bm q_4]^T$. $\bm Q$ is a diagonal weighting matrix. It is necessary to allow CoM z-direction location delta $\bm {\bm z}_{c}^{ref}-\bm {\bm z}_{c}$ to have certain flexibility in order to solve for the most optimal poses, in another word, we choose the weight in the z-direction location delta to be much smaller than that of x-direction location delta.

The optimization problem is subjected to several nonlinear constraints, shown in equation (\ref{eq:Constraint1}) to (\ref{eq:Constraint4}). The rim of each wheel is defined as the ground contact geometry whose geometry location $\bm p_{wheel}$ can be derived by FK with optimization variables $\mathbf X_i$. The rim of the wheel is constrained by equation(\ref{eq:Constraint1}) to be on the terrain in each pose.
In equation(\ref{eq:Constraint2}), we constrain x-direction of the rear wheel ground contact location $\bm p_{rw,x}$ to be less than that of rear hip $\bm p_{rh,x}$. Both of these locations can be derived by FK with $\mathbf X_i$. This will allow a larger support region in optimal poses, to prevent the robot from falling backward due to significantly large pitch angle during the task. Equation(\ref{eq:Constraint3}) can be implemented by integer programming in NLP. A custom function \texttt{InCollision} based on one point-line interception is applied here to determine whether the collision model is in contact with the terrain model (i.e. collision model should always be above terrain). The location of the collision model point cloud $\bm p_{cm}$ is determined by FK. Lastly, in equation(\ref{eq:Constraint4}), the joint angles $\mathbf q^*$ in the optimization variable are bounded by the physical joint limits of the hardware platform.

\begin{figure*}[!t]
     \center
     \begin{subfigure}[b]{1\textwidth}
         \centering
         \includegraphics[width=\textwidth]{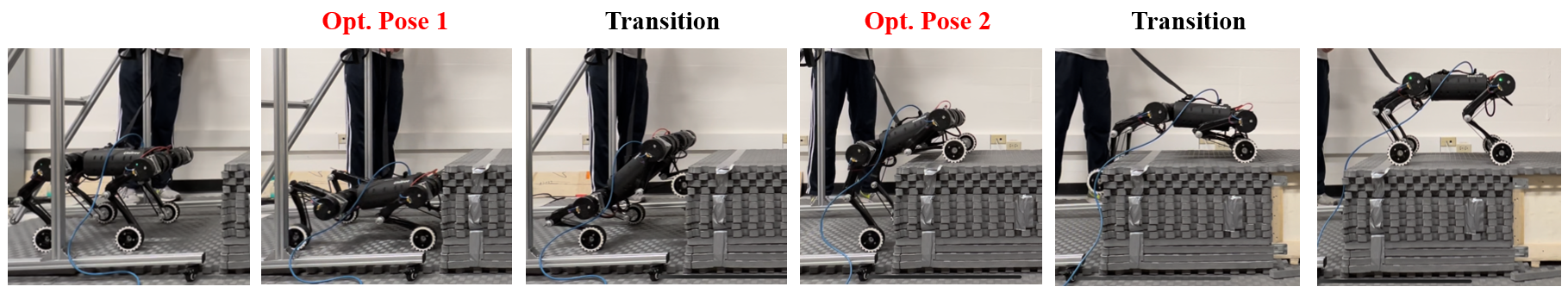}
         \caption{Hardware: Driving up one stair with a height of 0.36$\unit {m}$}
         \label{fig:upstair_hardware}
     \end{subfigure}
     
     \begin{subfigure}[b]{1\textwidth}
         \centering
         \includegraphics[width=\textwidth]{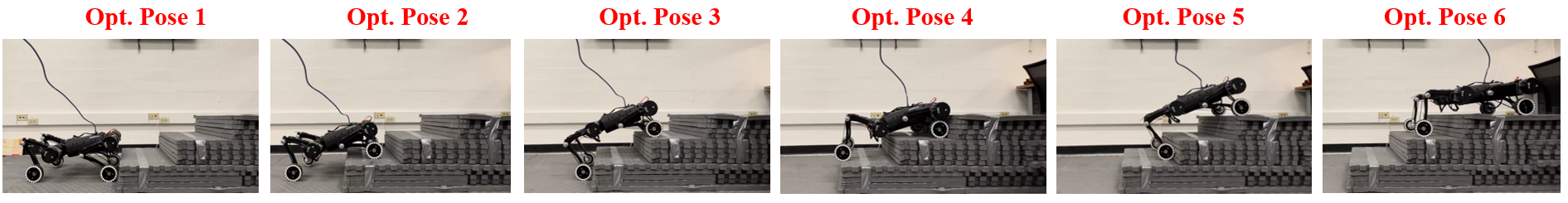}
         \caption{Hardware: Driving up 3 consecutive stairs with height of 0.125$\unit {m}$ and run of 0.30$\unit {m}$}
         \label{fig:3stair_hardware}
     \end{subfigure}
        \caption{{\bfseries Hardware Experiment Snapshots.} }
        \label{fig:simStair}
\end{figure*}
\subsection{Real-time Pose Planning}
\label{subsec:trackingController}

After the optimal poses are solved for a task, we use real-time pose planning in order to command desired joint angle and pitch angle online. The joint angle and pitch angle trajectory is linearly interpolated from initial pose to intermedian optimal poses and then to final pose. The general interpolation equation at time $t$ from pose $i$ ($\mathbf q_i $ and $ \bm \theta_i$) to pose $i+1$ ($\mathbf q_{i+1} $ and $ \bm \theta_{i+1}$) with a transition phase $\Delta t$ is as follows,
\begin{align}
\label{eq:interpolation}
\mathbf q_{des}(t) = \mathbf{q}_i + \frac{(\mathbf{q}_{i+1}-\mathbf{q}_i) \cdot 
(t-t_{0,i})}{\Delta t} \\
\bm \theta_{des}(t) = \bm \theta_i + \frac{(\bm \theta_{i+1}-\bm \theta_i) \cdot
(t-t_{0,i})}{\Delta t}
\end{align} 

Where $t_{0,i}$ is the initial time of transition from pose $i$ to pose $i+1$.
Since the optimization outputs only optimal joint and pitch angles, a tracking controller is needed to enable the robot to perform certain pose at desired location and timing. The optimal poses $\mathbf q^*$ are tracked by a joint PD controller, while the optimal pitch angle $\theta^*$ is tracked by QP balancing control. The timing of each pose $ \hat {\bm t}_1$ and $ \hat {\bm t}_2$ is estimated by the current average wheel joint speed $\overline{\dot{\bm q}}_{wheel}$ and terrain stair slope angle $\gamma$ and height $h$,
\begin{align}
\label{eq:tHat1}
{\hat {\bm t}_1} = \bm t + \frac{(\Delta \bm x - \bm R_{wheel}) }{\bm R_{wheel}\overline{\dot{\bm q}}_{wheel}}
\end{align} 
\begin{align}
\label{eq:tHat1}
{\hat {\bm t}_2} = {\hat {\bm t}_1}  + \frac{({\frac{h}{\sin(\gamma)}}+ \bm R_{wheel}) }{\bm R_{wheel}\overline{\dot{\bm q}}_{wheel}}
\end{align} 
$\bm t$ is the current timing at the start of estimation. 
\begin{align}
\label{eq:trackingControl}
\bm \tau = \bm \tau_{QP} + \bm K_{p,q}(\mathbf q^*-\mathbf q)+\bm K_{d,q}({\mathbf {\dot q}}^*-{\mathbf {\dot q}})
\end{align} 
The tracking controller works alongside QP force-based balance control to balance the robot while the robot is rolling over high obstacles. This approach is also a well-established tracking controller with motion planning (e.g. \cite{nguyen2019optimized}). The resulting control input $\bm \tau$ in terms of joint torques for thigh and calf joints is a combination of torque from balance controller $\bm \tau_{QP}$ and tracking controller, as shown in equation (\ref{eq:trackingControl}).

\section{Experiment Results}
\label{sec:Results}

In this section, we present highlighted hardware experiment results with the proposed approach.

The pose optimization framework can be implemented by many popular modern NLP solvers. We have implemented and executed the pose optimization in MATLAB \texttt{fmincon} Sequential Quadratic Programming (SQP) solver for our simulation and hardware experiments. The offline computation time for a single-stair pose optimization task is in the range of 0.3$\unit {s}$ to 0.5$\unit {s}$. As a benchmark, the PC hardware platform we use for offline motion planning has an AMD Ryzen 5-5600X CPU clocked at 4.65$\unit {GHz}$. We expect the computation cost to be scaled down further when the pose optimization is implemented in a C++ based solver such as IPOPT in the future.

\begin{figure}[!h]
    \center
	\includegraphics[width=1 \columnwidth]{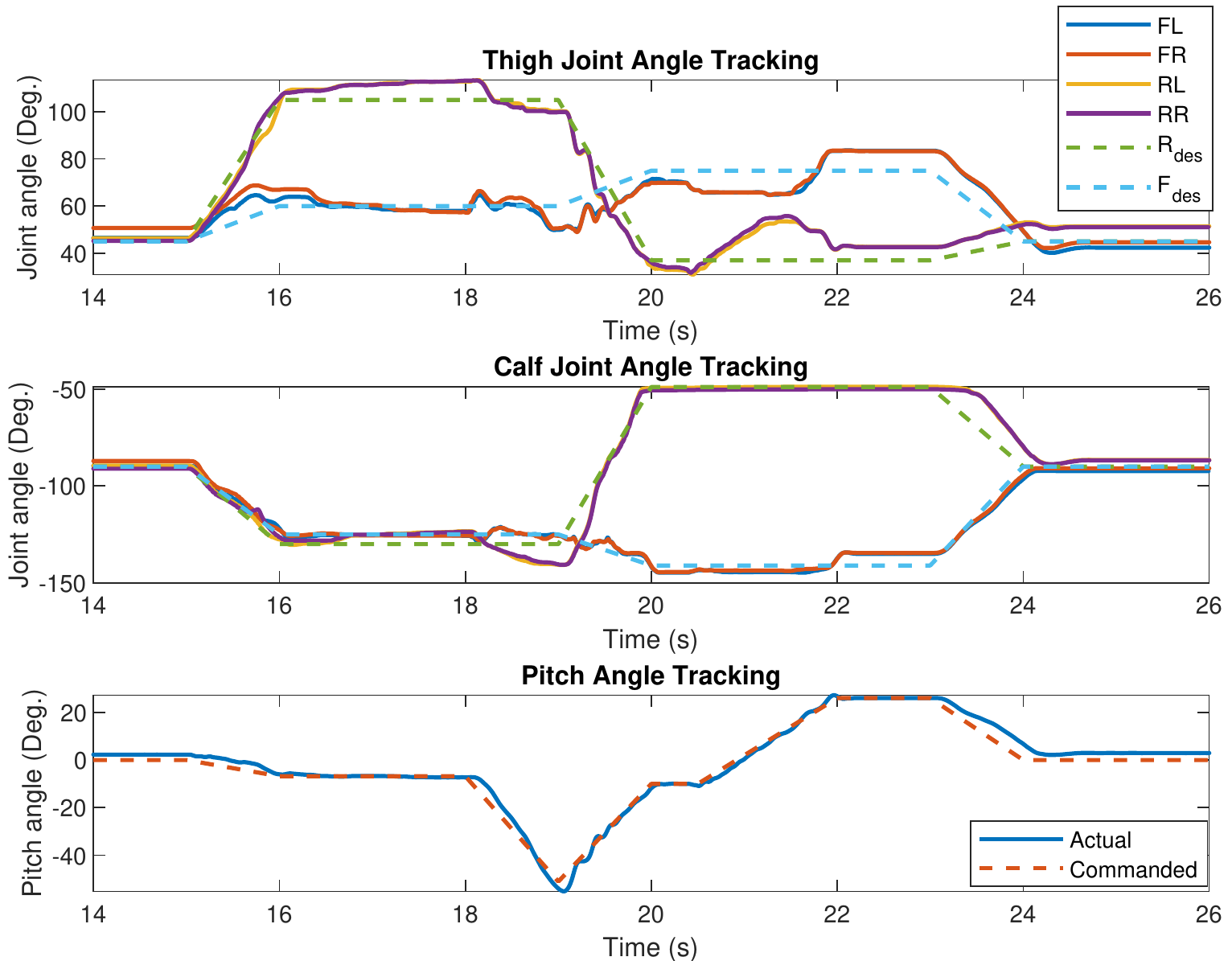}
    \caption{{\bfseries Hardware Experiment Plots.} Joint angle and pitch angle tracking plots in single-stair obstacle task with a height of 0.36$\unit {m}$, with pose optimization.}
        \label{fig:hardwarePlots}
\end{figure}

We have successfully validated our proposed approach on the real robot hardware.
In hardware experiments, our method with pose optimization has also shown its advantages compared to without pose optimization. 
Fig. \ref{fig:poseOptHardware} shows snapshots of experiment results comparing the performance of these two approaches. 
The robot is commanded to roll up a very simple 0.25$\unit{m}$ high ramp with a slope angle of ${30^{\circ}}$.
It is observed that using the nominal quadruped pose (without pose optimization), the CoM location falls towards the back of the leg support region and causes significant pitch angle error. In addition, the leg configuration does not favor traversing over slopes due to the collision of rear calf joints and the ground. Whereas, with our proposed approach using pose optimization, the CoM location stays centered at the support region and body pitch angle is minimal. 

We demonstrated our proposed approach in single-stair and 3-stair obstacle experiments. Our hardware successfully achieved climbing up a 0.36$\unit{m}$ stair. Nominal quadruped robot gait cannot climb up such a high stair because it is constrained by the nominal height of the robot. As a reference, the nominal standing height of our wheel-legged robot is 0.33$\unit {m}$. Snapshots of this experiment are shown in  Fig. \ref{fig:upstair_hardware}. 
The pitch and joint angle tracking plots of this successful single-stair experiment with pose optimization are shown in Fig. \ref{fig:hardwarePlots}. This also demonstrates the effectiveness of our balancing control presented in this paper.
We have also validated the versatility of our pose optimization framework in a 3-consecutive-stair task, snapshots are shown in Fig. \ref{fig:3stair_hardware}. Each stair has a stair rise of 0.125$\unit{m}$ and stair run of 0.30$\unit{m}$.



\section{Conclusions}
\label{sec:Conclusion}

In conclusion, we proposed an effective approach of balancing the 12 DoF wheel-legged robot with QP force-based control that employs modified simplified dynamics that considers the effects of wheel dynamics. By leveraging the wheel traction in high obstacle terrain locomotion, we have also proposed an optimization method in motion planning that solves for favorable poses during stair-climbing tasks. The optimal poses are tracked by a joint PD tracking controller, along with QP balancing controller. In hardware implementation, the robot is capable of climbing up on a ${ 0.36\:\unit{m}}$ stair (higher than robot nominal height). The versatility of the pose optimization framework is validated through successful multi-stair task experiments and proven to have superior performance as compared to normal quadruped poses during such tasks.
\balance
\bibliographystyle{IEEEtran}
\bibliography{ref}

\begin{thebibliography}{10}
\providecommand{\url}[1]{#1}
\csname url@samestyle\endcsname
\providecommand{\newblock}{\relax}
\providecommand{\bibinfo}[2]{#2}
\providecommand{\BIBentrySTDinterwordspacing}{\spaceskip=0pt\relax}
\providecommand{\BIBentryALTinterwordstretchfactor}{4}
\providecommand{\BIBentryALTinterwordspacing}{\spaceskip=\fontdimen2\font plus
\BIBentryALTinterwordstretchfactor\fontdimen3\font minus
  \fontdimen4\font\relax}
\providecommand{\BIBforeignlanguage}[2]{{%
\expandafter\ifx\csname l@#1\endcsname\relax
\typeout{** WARNING: IEEEtran.bst: No hyphenation pattern has been}%
\typeout{** loaded for the language `#1'. Using the pattern for}%
\typeout{** the default language instead.}%
\else
\language=\csname l@#1\endcsname
\fi
#2}}
\providecommand{\BIBdecl}{\relax}
\BIBdecl

\bibitem{kajita2008legged}
S.~Kajita and B.~Espiau, ``Legged robot,'' 2008.

\bibitem{nguyen2017dynamic}
Q.~Nguyen, A.~Agrawal, X.~Da, W.~C. Martin, H.~Geyer, J.~W. Grizzle, and
  K.~Sreenath, ``Dynamic walking on randomly-varying discrete terrain with
  one-step preview.'' in \emph{Robotics: Science and Systems}, vol.~2, no.~3,
  2017.

\bibitem{gong2019feedback}
Y.~Gong, R.~Hartley, X.~Da, A.~Hereid, O.~Harib, J.-K. Huang, and J.~Grizzle,
  ``Feedback control of a cassie bipedal robot: Walking, standing, and riding a
  segway,'' in \emph{2019 American Control Conference (ACC)}.\hskip 1em plus
  0.5em minus 0.4em\relax IEEE, 2019, pp. 4559--4566.

\bibitem{bledt2018cheetah}
G.~Bledt, M.~J. Powell, B.~Katz, J.~Di~Carlo, P.~M. Wensing, and S.~Kim, ``Mit
  cheetah 3: Design and control of a robust, dynamic quadruped robot,'' in
  \emph{2018 IEEE/RSJ International Conference on Intelligent Robots and
  Systems (IROS)}.\hskip 1em plus 0.5em minus 0.4em\relax IEEE, 2018, pp.
  2245--2252.

\bibitem{hutter2016anymal}
M.~Hutter, C.~Gehring, D.~Jud, A.~Lauber, C.~D. Bellicoso, V.~Tsounis,
  J.~Hwangbo, K.~Bodie, P.~Fankhauser, M.~Bloesch \emph{et~al.}, ``Anymal-a
  highly mobile and dynamic quadrupedal robot,'' in \emph{2016 IEEE/RSJ
  international conference on intelligent robots and systems (IROS)}.\hskip 1em
  plus 0.5em minus 0.4em\relax IEEE, 2016, pp. 38--44.

\bibitem{bouman2020autonomous}
A.~Bouman, M.~F. Ginting, N.~Alatur, M.~Palieri, D.~D. Fan, T.~Touma,
  T.~Pailevanian, S.-K. Kim, K.~Otsu, J.~Burdick \emph{et~al.}, ``Autonomous
  spot: Long-range autonomous exploration of extreme environments with legged
  locomotion,'' in \emph{2020 IEEE/RSJ International Conference on Intelligent
  Robots and Systems (IROS)}.\hskip 1em plus 0.5em minus 0.4em\relax IEEE,
  2020, pp. 2518--2525.

\bibitem{seok2014design}
S.~Seok, A.~Wang, M.~Y. Chuah, D.~J. Hyun, J.~Lee, D.~M. Otten, J.~H. Lang, and
  S.~Kim, ``Design principles for energy-efficient legged locomotion and
  implementation on the mit cheetah robot,'' \emph{Ieee/asme transactions on
  mechatronics}, vol.~20, no.~3, pp. 1117--1129, 2014.

\bibitem{pongas2007robust}
D.~Pongas, M.~Mistry, and S.~Schaal, ``A robust quadruped walking gait for
  traversing rough terrain,'' in \emph{Proceedings 2007 IEEE International
  Conference on Robotics and Automation}.\hskip 1em plus 0.5em minus
  0.4em\relax IEEE, 2007, pp. 1474--1479.

\bibitem{schwarz2016hybrid}
M.~Schwarz, T.~Rodehutskors, M.~Schreiber, and S.~Behnke, ``Hybrid
  driving-stepping locomotion with the wheeled-legged robot momaro,'' in
  \emph{2016 IEEE International Conference on Robotics and Automation
  (ICRA)}.\hskip 1em plus 0.5em minus 0.4em\relax IEEE, 2016, pp. 5589--5595.

\bibitem{grand2004stability}
C.~Grand, F.~Benamar, F.~Plumet, and P.~Bidaud, ``Stability and traction
  optimization of a reconfigurable wheel-legged robot,'' \emph{The
  International Journal of Robotics Research}, vol.~23, no. 10-11, pp.
  1041--1058, 2004.

\bibitem{grand2004decoupled}
C.~Grand, F.~BenAmar, F.~Plumet, and P.~Bidaud, ``Decoupled control of posture
  and trajectory of the hybrid wheel-legged robot hylos,'' in \emph{IEEE
  International Conference on Robotics and Automation, 2004. Proceedings.
  ICRA'04. 2004}, vol.~5.\hskip 1em plus 0.5em minus 0.4em\relax IEEE, 2004,
  pp. 5111--5116.

\bibitem{bjelonic2020whole}
M.~Bjelonic, R.~Grandia, O.~Harley, C.~Galliard, S.~Zimmermann, and M.~Hutter,
  ``Whole-body mpc and online gait sequence generation for wheeled-legged
  robots,'' \emph{arXiv preprint arXiv:2010.06322}, 2020.

\bibitem{klemm2020lqr}
V.~Klemm, A.~Morra, L.~Gulich, D.~Mannhart, D.~Rohr, M.~Kamel, Y.~de~Viragh,
  and R.~Siegwart, ``Lqr-assisted whole-body control of a wheeled bipedal robot
  with kinematic loops,'' \emph{IEEE Robotics and Automation Letters}, vol.~5,
  no.~2, pp. 3745--3752, 2020.

\bibitem{nguyen2019optimized}
Q.~Nguyen, M.~J. Powell, B.~Katz, J.~Di~Carlo, and S.~Kim, ``Optimized jumping
  on the mit cheetah 3 robot,'' in \emph{2019 International Conference on
  Robotics and Automation (ICRA)}.\hskip 1em plus 0.5em minus 0.4em\relax IEEE,
  2019, pp. 7448--7454.

\bibitem{li2021force}
J.~Li and Q.~Nguyen, ``Force-and-moment-based model predictive control for
  achieving highly dynamic locomotion on bipedal robots,'' \emph{arXiv preprint
  arXiv:2104.00065}, 2021.

\bibitem{katz2019mini}
B.~Katz, J.~Di~Carlo, and S.~Kim, ``Mini cheetah: A platform for pushing the
  limits of dynamic quadruped control,'' in \emph{2019 international conference
  on robotics and automation (ICRA)}.\hskip 1em plus 0.5em minus 0.4em\relax
  IEEE, 2019, pp. 6295--6301.

\bibitem{bellegarda2019trajectory}
G.~Bellegarda and K.~Byl, ``Trajectory optimization for a wheel-legged system
  for dynamic maneuvers that allow for wheel slip,'' in \emph{2019 IEEE 58th
  Conference on Decision and Control (CDC)}.\hskip 1em plus 0.5em minus
  0.4em\relax IEEE, 2019, pp. 7776--7781.

\bibitem{winkler2018gait}
A.~W. Winkler, C.~D. Bellicoso, M.~Hutter, and J.~Buchli, ``Gait and trajectory
  optimization for legged systems through phase-based end-effector
  parameterization,'' \emph{IEEE Robotics and Automation Letters}, vol.~3,
  no.~3, pp. 1560--1567, 2018.

\bibitem{luo2014design}
Y.~Luo, Q.~Li, and Z.~Liu, ``Design and optimization of wheel-legged robot:
  Rolling-wolf,'' \emph{Chinese journal of mechanical engineering}, vol.~27,
  no.~6, pp. 1133--1142, 2014.

\bibitem{medeiros2020trajectory}
V.~S. Medeiros, E.~Jelavic, M.~Bjelonic, R.~Siegwart, M.~A. Meggiolaro, and
  M.~Hutter, ``Trajectory optimization for wheeled-legged quadrupedal robots
  driving in challenging terrain,'' \emph{IEEE Robotics and Automation
  Letters}, vol.~5, no.~3, pp. 4172--4179, 2020.

\bibitem{chignoli2021humanoid}
M.~Chignoli, D.~Kim, E.~Stanger-Jones, and S.~Kim, ``The mit humanoid robot:
  Design, motion planning, and control for acrobatic behaviors,'' in \emph{2020
  IEEE-RAS 20th International Conference on Humanoid Robots (Humanoids)}.\hskip
  1em plus 0.5em minus 0.4em\relax IEEE, 2021, pp. 1--8.

\bibitem{besseron2005locomotion}
G.~Besseron, C.~Grand, F.~B. Amar, F.~Plumet, and P.~Bidaud, ``Locomotion modes
  of an hybrid wheel-legged robot,'' in \emph{Climbing and Walking
  Robots}.\hskip 1em plus 0.5em minus 0.4em\relax Springer, 2005, pp. 825--833.

\bibitem{amar2004performance}
F.~B. Amar, C.~Grand, G.~Besseron, and F.~Plumet, ``Performance evaluation of
  locomotion modes of an hybrid wheel-legged robot for self-adaptation to
  ground conditions,'' in \emph{ASTRA'04, 8th ESA Workshop on Advanced Space
  Technologies for Robotics and Automation}, 2004.

\bibitem{ni2020posture}
L.~Ni, F.~Ma, and L.~Wu, ``Posture control of a four-wheel-legged robot with a
  suspension system,'' \emph{IEEE Access}, vol.~8, pp. 152\,790--152\,804,
  2020.

\bibitem{focchi2017high}
M.~Focchi, A.~Del~Prete, I.~Havoutis, R.~Featherstone, D.~G. Caldwell, and
  C.~Semini, ``High-slope terrain locomotion for torque-controlled quadruped
  robots,'' \emph{Autonomous Robots}, vol.~41, no.~1, pp. 259--272, 2017.

\end{thebibliography}

\end{document}